\documentclass[twocolumn,11pt]{article}
\usepackage[utf8]{inputenc}
\DeclareUnicodeCharacter{00A0}{}
\DeclareUnicodeCharacter{00A0}{~}
\usepackage{lmodern}
\usepackage{mathtools, cuted}
\usepackage{titlesec}
\usepackage{blindtext}
\usepackage{rotating}
\usepackage{varwidth} 
\newlength\titleindent
\usepackage[top=2cm, bottom=3cm, left=1.5cm, right=1.5cm]{geometry}

\usepackage{import}
\usepackage{graphicx}
\usepackage{color}
\usepackage{natbib}
\usepackage[table,xcdraw,svgnames]{xcolor}
\usepackage{tabularx,booktabs}
\usepackage{pgfplots}
\usepackage{pgfplotstable}
\usepackage{subcaption}
\usepackage{graphicx}
\usepackage{multirow}
\usepackage{mathtools}
\usepackage{algorithm, algorithmic}
\usepackage{makeidx}
\usepackage{soul}
\usepackage{xcolor}
\usepackage{amsthm,amssymb,amsmath}
\usepackage{pifont}
\setcitestyle{open={[},authoryear,close={]}}
\usepackage{tabularx}
\usepackage{lscape}
\usepackage{pdflscape}
\usepackage{multirow}
\usepackage{tabulary}
\usepackage{supertabular,booktabs}
\usepackage{ragged2e}
\newcolumntype{L}[1]{>{\raggedright\arraybackslash}p{#1}}
\newcolumntype{C}[1]{>{\centering\arraybackslash}p{#1}}
\newcolumntype{R}[1]{>{\raggedleft\arraybackslash}p{#1}}
\newcolumntype{M}[1]{>{\centering\arraybackslash}m{#1}}
\usepackage{fourier, heuristica}
\usepackage{array, booktabs}
\usepackage{graphicx}
\usepackage{amsmath}
\usepackage{supertabular}
\usepackage{caption}

\usepackage[font={small,it}]{caption}
\usepackage{ltablex}
\DeclareCaptionFont{gray}{\color{gray}}

\usepackage{amsmath}

\usepackage{amssymb}
\usepackage{latexsym}

\usepackage{url}
\AtBeginDocument{%
  \hypersetup{
    citecolor=blue!60!black,
    linkcolor=blue!60!black,   
    urlcolor=blue!60!black}}

\usepackage{tikz}
\usetikzlibrary{arrows.meta}
\usetikzlibrary{through}
\usetikzlibrary{matrix}
\usetikzlibrary{decorations.pathreplacing}
\tikzset{%
	>={Latex[width=2mm,length=2mm]},
	base/.style = {rectangle, rounded corners, draw=black,
		minimum width=4cm, minimum height=1cm,
		text centered, font=\sffamily},
	new/.style = {rectangle, draw=black,
		minimum width=4cm, minimum height=1cm,
		text centered, font=\sffamily},
	activityStarts/.style = {base, fill=cyan!90!blue},
	startstop/.style = {base, fill=red!40},
	activityRuns/.style = {base, fill=green!70!black},
	rec/.style = {new, fill=white},
	process/.style = {base, minimum width=2.5cm, fill=orange!15,
		font=\ttfamily},
	data/.style = {base, minimum width=2.5cm, fill=cyan!80!blue,
		font=\ttfamily},
}
\usetikzlibrary{mindmap,shadows}

\usepackage[hidelinks, pdfencoding=auto,colorlinks]{hyperref}
\title{
{\Large \bfseries Multilingual Natural Language Processing Model for Radiology Reports
\\ \textit{The Summary is all you need!}
}
}

\author{Mariana Lindo$^{1,2}$ \and Ana Sofia Santos$^{1,2}$ \and André Ferreira$^{1,2}$ \and Jianning Li$^{2}$ \and Gijs Luijten$^{2,4}$ \and Gustavo Correia$^{1}$ \and Moon Kim$^{2}$ \and Benedikt Michael Schaarschmidt$^{3}$ \and Cornelius Deuschl$^{3}$ \and Johannes Haubold$^{2,3}$ \and Jens Kleesiek$^{2}$ \and Jan Egger$^{2}$ \and Victor Alves$^1$}

\date{$^1$\small{ALGORITMI Research Centre/LASI, University of Minho, Braga, Portugal} \\
$^2$\small{Institute for Artificial Intelligence in Medicine, University Medicine Essen (AöR), Essen, Germany} \\
$^3$\small{Department of Diagnostic and Interventional Radiology and Neuroradiology, University Medicine Essen (AöR), Essen, Germany}\\
$^4$\small{Institute for Computer Graphics and Vision,  Graz University of Technology, Graz, Austria}\\
}

\begin{document}
\twocolumn[\maketitle \begin{@twocolumnfalse} 
\hrule
\begin{abstract}
\large
The impression section of a radiology report summarizes important radiology findings and plays a critical role in communicating these findings to physicians. However, the preparation of these summaries is time-consuming and error-prone for radiologists. Recently, numerous models for radiology report summarization have been developed. Nevertheless, there is currently no model that can summarize these reports in multiple languages. Such a model could greatly improve future research and the development of Deep Learning models that incorporate data from patients with different ethnic backgrounds. In this study, the generation of radiology impressions in different languages was automated by fine-tuning a model, publicly available, based on a multilingual text-to-text Transformer to summarize findings available in English, Portuguese, and German radiology reports. In a blind test, two board-certified radiologists indicated that for at least 70\% of the system-generated summaries, the quality matched or exceeded the corresponding human-written summaries, suggesting substantial clinical reliability. Furthermore, this study showed that the multilingual model outperformed other models that specialized in summarizing radiology reports in only one language, as well as models that were not specifically designed for summarizing radiology reports, such as ChatGPT.
\\ \leavevmode \\


\hrule
\end{abstract}

\end{@twocolumnfalse}] 
\vspace{2.5cm}
    \vspace{5cm}
	\section*{Introduction}
\vspace{-2mm}

\label{sec:Introduction}
Medical imaging techniques like X-rays and Computed Tomography (CT) are widely employed for disease diagnosis, treatment planning, and guidance \cite{hill2001medical}. In routine clinical practice, these imaging studies are conveyed through a written radiology report (RR), which serves as the official record documenting diagnostic, interventional, or therapeutic examinations or procedures utilizing medical imaging data \cite{liang2022fine}. Within the standard clinical workflow, a radiologist initially records detailed findings in the report and subsequently provides a summary of the most significant observations. The summary, also known as the impression, holds paramount importance in an RR, as referring physicians focus their attention on it when reviewing the report \cite{bosmans2011radiology}. However, radiologists often encounter difficulties in accurately documenting critical findings in the summary, which may result in forgetting information or omitting findings they consider less important. Therefore, report writing is prone to errors and can lead to misunderstandings. In addition, writing the summary is time-consuming and quite tedious, and dictating the findings can be highly repetitive \cite{weber2020improving}.

Recently, the rapid advancements in Deep Learning (DL) techniques have paved the way for the introduction of summary models in the field of Natural Language Processing (NLP) \cite{narayan2020stepwise}. However, to the best of our knowledge, no multilingual summary models have been developed specifically for summarizing RRs across different languages. Such a model could greatly facilitate future research, utilization of Data Mining techniques, and the development of DL models aiming to incorporate data from patients of diverse ethnic backgrounds. Furthermore, a summarization model could alleviate the stress and workload on radiologists by automatically generating succinct summaries that retain essential information, thus enhancing communication between radiologists and referring physicians \cite{kahn2009toward}. The primary objective of this project is to develop a multilingual NLP model dedicated to summarizing RRs in various languages.
    \section{Related Work}
\label{sec:Related_Work}

The study by \cite{zhang2018learning} was one of the pioneers in developing models for summarizing RRs. They automated the generation of impressions by using a Recurrent Neural Network-based pointer generator \cite{vinyals2015pointer}, which was responsible for the summary of the findings, and a Bidirectional Long Short-Term Memory encoder \cite{zhang2015bidirectional}, which was used to separately encode background information and findings of an RR. After training and testing the model with 60,990 and 17,425 RRs, respectively, the results showed that the model outperformed the existing non-neuronal and neuronal baselines under the ROUGE metrics \cite{lin-2004-rouge}. In addition, experienced radiologists confirmed that 67\% of the system summaries tested were at least as good as the corresponding human-written summaries.

\cite{cai2021chestxraybert} developed a pre-trained language model in thoracic radiology, called ChestXRay Bidirectional Encoder Representations from Transformers (ChestXRay BERT), to automatically summarize RRs. They started by collecting 85 radiology-related scientific articles from PubMed Central \cite{roberts2001pubmed} and pre-trained the ChestXRayBERT on them. Then, a Transformer decoder was added to ChestXRayBERT to create an abstract summary model, which was fine-tuned using RRs. For training and evaluation of model performance, the Indiana University Chest X-Ray (IU X-ray) \cite{demner2016preparing} and MIMIC Chest X-ray (MIMIC-CXR) \cite{johnson2019mimic} datasets were combined and then divided into training/validation and test sets. The results showed that ChestXRayBERT achieved significant improvement compared with other neural network-based abstract summary models.

    \section{Model Architecture}
\label{sec:Model_Architecture}

To develop a model capable of summarizing RRs in multiple languages, the Multilingual Text-to-Text Transfer Transformer (mT5) \cite{xue2020mt5}, which can be accessed on the Hugging Face Hub, was used as a starting point. The mT5 model represents a multilingual variant of the Text-to-Text Transfer Transformer (T5) \cite{raffel2020exploring}, which is a pre-trained language model specifically designed for a broad range of text-based NLP tasks, including abstractive summarization and translation. The T5 model employs a uniform Sequence-to-sequence (Seq2seq) structure to address these NLP challenges \cite{chen2018fast}. The T5 architecture shares similarities with the original Transformer model \cite{vaswani2017attention}. However, there are a few distinctions: the layer normalization operates outside the residual path, biases have been eliminated, and a different positional embedding scheme is used.

The mT5 architecture and training procedure are very similar to those of T5. In particular, mT5 is based on the T5.1.1 checkpoint \cite{t5}, which improves upon T5 by using Gaussian Error Gated Linear Units nonlinearities \cite{shazeer2020glu}, scaling the output dimensionality of the Feed-Forward Neural Network, other sublayers, and embeddings in the larger models. Furthermore, mT5 is pretrained exclusively on unlabeled data without utilizing dropout. The pretraining dataset for mT5 is mC4, a multilingual variant of the Colossal Clean Crawled Corpus dataset. This dataset encompasses natural text in 101 languages gathered from the publicly available Common Crawl Web Scrape. The mT5 model offers five model sizes, one of which is the base size with 580M parameters, and was used in this project. One difference between T5 and mT5 is that mT5 was pre-trained on mC4 only, without any supervised training. As a result, this model needs to be fine-tuned before it can be used for a downstream task \cite{xue2020mt5}. 
    
\section{Material and Methods}
\label{sec:Material_and_Methods}
This study was approved by the ethics committee (22-10997-BO) and only fully anonymized data have been used.

To train the multilingual final model, we started a local open-source network and used the processing pipeline shown in Figure 1, which illustrates the use of 5 datasets and 7 models. Each dataset is formally denoted as \emph{D$^{n}_{l}$}, where \emph{n} is the name of the dataset and \emph{l} is the language in which the dataset was used. The value of \emph{l} can be EN (English), PT (Portuguese), and/or GER (German). Similarly, each model is formally represented as \emph{M$^{p}_{tl}$}, where \emph{p} is the purpose for which the model was developed, and \emph{tl} is the target language(s) for which the model was trained. The value of \emph{p} can be \emph{base} if the model used was the base version; \emph{summaries} if the model was trained for summary texts; \emph{rr-1000} if the model was tuned to summarize RRs with a Max New Tokens Parameter (MNTP) of 1,000; and \emph{translation} if the model was trained specifically to translate RRs.

Since the mT5 model was not specifically fine-tuned for a downstream task, the initial step involved fine-tuning the model for English text summarization. Only the English instances from the Multilingual Amazon Reviews Corpus (MARC) dataset were employed, resulting in 125,893, 3,166, and 3,146 reviews for the training, validation, and test splits, respectively, in the \emph{D$^{MARC}_{EN}$} dataset. The \emph{review\_body} columns served as inputs, while the \emph{review\_title} columns were used as targets. The optimizer AdamW and a linear decay learning rate scheduler, decreasing from the maximum value of 2e$^{-5}$ to 0, were employed for training. To ensure concise review summaries, the MNTP was set to 50 tokens. The fine-tuned model checkpoint derived from the \emph{M$^{base}_{EN}$} model for English text summarization was defined as \emph{M$^{summaries}_{EN}$}, trained for 10 epochs with a Batch Size (BS) of 8. The workstation used has 64 GB of RAM, the CPU is an Intel Xeon E5-1650, and the GPU is an NVIDIA P6000 with 24 GB of memory.

Subsequently, the \emph{M$^{summaries}_{EN}$} model was further fine-tuned for summarizing English RRs using the MIMIC-CXR dataset. The RRs in this dataset consisted of semi-structured text, with various sections such as patient history, findings, and impressions. To address imbalanced data issues, some of the most frequent impressions were removed, ensuring that one impression occurred less than 2\% of the time. The balanced dataset, \emph{D$^{MIMIC-CXR}_{EN}$}, comprised 5,816, 1,455, and 1,818 instances for training, validation, and testing, respectively. The resulting model, \emph{M$^{rr-1000}_{EN}$} was trained with an MNTP value of 1,000 and a BS of 1, due to the lack of graphical capacity of the workstation used.

To develop a model capable of summarizing RRs in other language, starting with Portuguese, reports in Portuguese were introduced.  Initially, RRs from a Portuguese hospital were employed, but this led to overfitting as most of the reports were identical. Consequently, the generated summaries became repetitive, irrespective of the input findings. To overcome this issue, an alternative approach was adopted, which involved translating the \emph{D$^{MIMIC-CXR}_{EN}$} dataset into Portuguese. This resulted in the creation of a new dataset defined as \emph{D$^{MIMIC-CXR}_{PT}$}, comprising 10,369, 2,593, and 3,241 instances for training, validation, and testing, respectively. The translation of the \emph{D$^{MIMIC-CXR}_{EN}$} dataset was performed using the GoogleTranslator library. However, this process proved to be time-consuming and unsuitable for private datasets due to patient confidentiality concerns. Moreover, nowadays, with the increasing international mobility of people, it is becoming more and more attractive to have specialized and automatic translations of RRs. Thus, an alternative method was sought to translate the RRs. Considering that the mT5 model is proficient in Seq2seq tasks, including translation, it was fine-tuned to specifically translate impressions and findings from English to Portuguese. The \emph{D$^{MIMIC-CXR}_{PT}$} dataset served as the input to develop the \emph{M$^{translation}_{EN-PT}$} model, which was trained for 20 epochs to reach stability, with BS 1 and MNTP 1,000.

With the \emph{M$^{translation}_{EN-PT}$} model proficient in translating English to Portuguese impressions and findings, the same sections of the English public IU X-Ray dataset were translated. However, the translated dataset exhibited imbalanced data, necessitating the implementation of a solution. To address this issue, the dataset was balanced, resulting in the creation of the \emph{D$^{IU X-Ray}_{PT}$} dataset, which comprised 1,273, 319, and 399 instances for training, validation, and testing, respectively. Using this dataset, the \emph{M$^{rr-1000}_{PT}$} model was developed and fine-tuned specifically for summarizing Portuguese RRs. The training process lasted for 10 epochs to achieve stability, with a BS of 1 and an MNTP of 1,000 being employed.

The next step was to train a model capable of summarizing RRs in German. For this purpose, RRs related to CTs from a private dataset were utilized to construct the \emph{D$^{German RRs}_{GE}$}. This dataset consisted of 34,166, 8,542, and 10,677 instances allocated for training, validation, and testing, respectively. Using the German RRs dataset, the \emph{M$^{rr-1000}_{GE}$} model was developed and fine-tuned specifically for summarizing German RRs. The training process was carried out for 17 epochs to ensure stability, employing a BS of 1 and setting the MNTP to 1,000.

Up until now, various summary models have been created to summarize radiology reports in English, Portuguese, and German individually. However, these existing models do not align with the objective of this project, which aims to develop a single model capable of summarizing reports in multiple languages. To achieve this goal, a final step was undertaken, involving the training of a multilingual model, the \emph{M$^{rr-1000}_{EN, PT, GE}$}. To do so, radiology reports from the datasets \emph{D$^{MIMIC-CXR}_{EN}$}, \emph{D$^{IU X-Ray}_{PT}$}, and \emph{D$^{German RRs}_{GE}$} were utilized. Nonetheless, the number of German reports exceeded those in English by approximately 6 times, and English reports were around 5 times more abundant than Portuguese ones. Training the multilingual model with reports in the same ratio as their occurrence would lead to overfitting towards generating German summaries, given the significantly higher number of German instances. To counter this, only as many reports as those in Portuguese (1,991) were used for training the multilingual model in each language. The training, testing, and validation splits consisted of 1,591, 200, and 200 instances, respectively, for each language.

The \emph{M$^{rr-1000}_{GE}$} model was chosen as the checkpoint model for fine-tuning due to the favorable language transfer learning abilities of mT5-based models. Language transfer learning involves leveraging knowledge from one language to enhance performance in another. The mT5 model is specifically designed to comprehend shared representations across multiple languages during its pre-training phase. This facilitates the model's understanding of the underlying structure and common patterns across different languages, making the \emph{M$^{rr-1000}_{EN, PT, GE}$} model suitable for summarizing RRs in English, Portuguese, and German after fine-tuning.

\begin{figure}[h!]
  \centering
  \includegraphics[width=1\columnwidth]{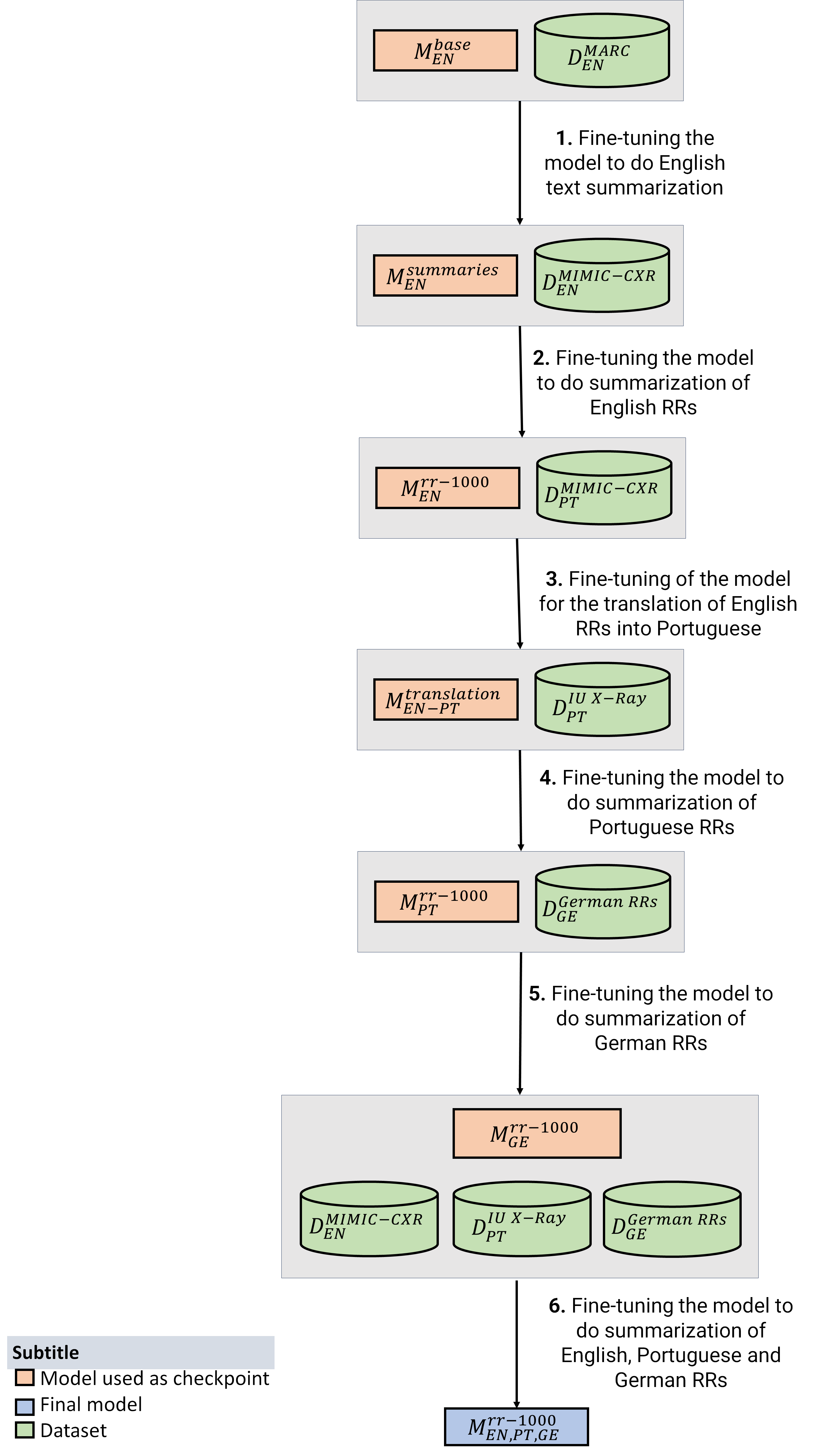}
  \caption{Processing steps followed to train the final model.}
  \label{Fig1}
\end{figure}%

    \section{Results}
\label{sec:Results}

Different variants of the ROUGE metric were used to quantitatively evaluate the results obtained with the different summarization models, as this is the most commonly used metric to evaluate summaries. Table 1 summarizes the results. 

A potential shortcoming of the ROUGE metrics is that they only measure the similarity between the Generated Summary (GS) and the Reference Summary (RS) but do not adequately reflect the overall grammar or utility of the predictions. Therefore, evaluations were also performed with a radiologist to understand the clinical validity of the summaries generated by the models. In this evaluation, 30 examples were randomly selected from the test sets. All models were applied to these examples, and the GSs were presented to the radiologist along with the corresponding human-written RSs. The radiologist was asked to decide which of the summaries was better, or if they were of roughly the same quality. The expert was also asked to rate the GSs on a scale of 5 (very good) to 1 (very poor) in terms of Readability (R), Factual Correctness, Completeness (FCC), and Overall Quality (OQ). The average results of these metrics are shown in Table 2, as well as the percentage of GSs that were rated as better than or equal to the RSs. 

In addition to the qualitative analysis performed by the radiologist, a second analysis was performed by presenting ChatGPT (\cite{shahriar2023let, li2023chatgpt}) with sections of findings from various reports and asking it to provide a summary of them. Some examples of  summaries obtained by the ChatGPT and by the models are shown in Table 3.

    \section{Discussion and Conclusion}
\label{sec:Conclusion_and_Discussion}

Based on the analysis of the results presented in Table 1, it can be concluded that the ROUGE values obtained fall within the range reported in existing literature for the task of summarizing radiology reports \cite{cai2021chestxraybert}. By examining Table 2, it becomes apparent that the multilingual \emph{M$^{rr-1000}_{EN, PT, GE}$} generally has the best qualitative results compared to the other checkpoint models that summarize radiology reports in only one language. While the \emph{M$^{rr-1000}_{EN, PT, GE}$} produced summaries in English with slightly less factual accuracy, completeness, and overall quality than the \emph{M$^{rr-1000}_{EN}$} and summaries in Portuguese with slightly less readability than the \emph{M$^{rr-1000}_{PT}$}, the summaries produced by the multilingual model in German were better than the summaries produced by the \emph{M$^{rr-1000}_{GE}$} in all metrics. Moreover, the percentage of summaries generated that are equal to or better than the reference summaries is always higher for the \emph{M$^{rr-1000}_{EN, PT, GE}$} than for the best model summarizing RRs in English, Portuguese, or German. This indicates that GSs produced by the multilingual model are more realistic and, in most cases, better than those produced by a radiologist. It is also important to note that all values for the metrics of readability, factual correctness and completeness, and overall quality are between 4 and 5, i.e., from good to very good, while some of the same values for the monolingual summarization models have lower values than 4.

Training a single model such as \emph{M$^{rr-1000}_{EN, PT, GE}$} to summarize radiology reports in multiple languages produces better results than training separate models for each language. One of the reasons that could lead to better performance is that the \emph{M$^{base}_{EN}$}, which served as the basis for developing the final \emph{M$^{rr-1000}_{EN, PT, GE}$}, has shared representations for all languages. This means that it can learn multilingual embeddings by leveraging the similarities and common structures in different languages, allowing it to better generalize across languages. By jointly fine-tuning on English, Portuguese, and German radiology reports, the model can capture cross-linguistic information, which improves its ability to produce summaries in all three languages. In addition, the model is also able to apply language transfer learning in the summarization task, meaning that knowledge acquired in one language can improve performance in another. For example, the model can learn effective summarization strategies in English, and some of this knowledge can be transferred to Portuguese and German, resulting in improved performance in those languages as well.

The information presented in Table 3 further supports the ability of the developed models to summarize RRs. The results demonstrate that the generated summaries are more similar to the original summaries, which is expected given that ChatGPT is not specialized in summarizing RRs. It was also observed that ChatGPT produces summaries that are either the same length or longer than the original reports. Consequently, it can be concluded that ChatGPT does not synthesize information but rather rephrases the information contained in the findings, lacking the capability to discern the relative importance of different pieces of information. Therefore, these results confirm that employing specialized models for summarizing RRs is more effective than using a nonspecialized model for this task.

In conclusion, this research resulted in the development of the \emph{M$^{rr-1000}_{EN, PT, GE}$} model, publicly available \footnote[1]{Available at \url{https://huggingface.co/MarianaLC/mt5-rr-1000}.}, which can generate summaries of RRs in English, Portuguese and German that are as good as or even superior to those produced by radiologists. The favorable outcomes suggest that the GSs can be utilized in conjunction with medical imaging and other data to develop multimodal DL models, such as those used for diagnosis, prediction, or treatment, without relying solely on data from patients of a single ethnicity. Moreover, the generated summaries can alleviate the workload of radiologists, who would only need to validate the accuracy of the generated summaries and provide additional information if necessary, rather than creating summaries from scratch. The radiologist that evaluated the English and Portuguese summaries even expressed a preference for this approach, stating, "I would rather do it this way than write a summary from scratch!".
Future work could see an assessment of the RECIST (Response Evaluation Criteria in Solid Tumours) rules from the summary, e.g. when is it a progressive disease (PD), stable disease (SD), stable response (SR) or treatment response (TR).

\begin{@twocolumnfalse} 
\begin{table}[!h]
\begin{small}
\centering
\caption{\centering Quantitative results of the summarization models.}
\begin{tabularx}{\linewidth}{|M{3.3cm}|M{3.5cm}|M{3.3cm}|M{3.3cm}|M{3.3cm}|}
\hline
\textbf{Model Name} & \textbf{ROUGE-1} & \textbf{ROUGE-2} & \textbf{ROUGE-L} & \textbf{ROUGE L-sum} \\ \hline

\emph{M$^{summaries}_{EN}$} & 26.33 & 16.94 & 25.61 & 25.68 \\ \hline
\emph{M$^{rr-1000}_{EN}$} & 40.56 & 26.96 & 37.72 & 39.04 \\ \hline
\emph{M$^{rr-1000}_{PT}$} & 42.48 & \textbf{32.39} & 41.27 & 41.71 \\ \hline
\emph{M$^{rr-1000}_{GE}$} & 41.50 & 28.08 & 38.26 & 40.37 \\ \hline
\emph{M$^{rr-1000}_{EN, PT, GE}$} & \textbf{46.11} & 32.31 & \textbf{43.54} & \textbf{44.93} \\ \hline
\end{tabularx}
\end{small}
\end{table}

\begin{table}[!h]
\begin{small}
\renewcommand\thetable{\arabic{table}}
\setcounter{table}{1}
\caption{Results of the qualitative radiologist evaluation.}
\begin{tabularx}{\linewidth}{|M{2.7cm}|M{2.7cm}|M{2.7cm}|M{2.6cm}|M{2.7cm}|M{2.7cm}|}
\hline
\textbf{Model Name} & \textbf{Language} & \textbf{GS $\geq$ RS (\%)} & \textbf{R (1-5)} & \textbf{FCC (1-5)} & \textbf{OQ (1-5)} \\ \hline
\emph{M$^{rr-1000}_{EN}$} & \multirow{3}{*}{-} & 66.67 & 4.73 & 4.10 & 4.20 \\ \cline{1-1} \cline{3-6} 
\emph{M$^{rr-1000}_{PT}$} &                    & 63.33 & 4.80 & 4.03 & 3.93 \\ \cline{1-1} \cline{3-6} 
\emph{M$^{rr-1000}_{GE}$} &                    & 50.00 & 4.33 & 4.20 & 4.00 \\ \hline
\multirow{3}{*}{\emph{M$^{rr-1000}_{EN, PT, GE}$}} & English            & 80.00 & 4.73 & 4.07 & 4.07 \\ \cline{2-6} 
                   & Portuguese         & 73.33 & 4.70 & 4.13 & 4.10 \\ \cline{2-6} 
                   & German             & \textbf{93.33} & \textbf{4.90} & \textbf{4.23} & \textbf{4.40} \\ \hline
\end{tabularx}
\end{small}
\end{table}
\end{@twocolumnfalse}

\begin{@twocolumnfalse} 
\setcounter{table}{2}
\begin{landscape}

\begin{center}
\small
\begin{longtable}{|p{6.5cm}|p{2.5cm}|p{7.5cm}|p{3.5cm}|}
\caption{Results of the qualitative radiologist evaluation of the summarization models.} \\
\hline
& \multicolumn{3}{|p{9.5cm}|}{\textbf{Summary}} \\ \hline
\textbf{Findings} & \textbf{Original} & \textbf{ChatGPT} & \textbf{Model} \\ \hline
\endfirsthead

\multicolumn{4}{c}%
{{\tablename\ \thetable{2} -- Continued from previous page}} \\
\hline
\textbf{Findings} & \textbf{Original} & \textbf{ChatGPT} & \textbf{Model} \\ \hline
\endhead

\hline \multicolumn{4}{|r|}{{Continued on next page}} \\ \hline
\endfoot

\hline
\endlastfoot

PA and lateral views of the chest provided. Subtle opacity in the upper lungs may represent an early pneumonia. The hila appear somewhat prominent which may indicate mild vascular congestion. No overt edema or large effusion is seen. Dense atherosclerotic calcification along the aorta is noted. The heart is top-normal in size. Severe degenerative disease at the right shoulder is again seen. There is a dextroscoliosis of the T-spine. & Subtle opacities in the upper lungs is concerning for an early pneumonia. Mild pulmonary vascular congestion also noted. & The chest X-ray includes PA and lateral views. There is a subtle opacity in the upper lungs, indicating a possible early pneumonia. The hila (structures near the lungs' root) appear slightly prominent, suggesting mild vascular congestion. No significant edema or large effusion is observed. Dense atherosclerotic calcification is noted along the aorta. The heart size is within normal limits. Severe degenerative disease is visible in the right shoulder, and there is a dextroscoliosis of the thoracic spine. & Subtle opacity in the upper lungs may represent an early pneumonia. Mild vascular congestion. \\ \hline

Contornos mediastinais de tamanho normal do coração. Atelectasia subsegmentar versus cicatriz no meio do pulmão direito e no lobo inferior esquerdo. Sem doença focal do espaço aéreo. Sem derrame pleural ou pneumotórax. Baixos volumes pulmonares. As estruturas ósseas visualizadas não têm aparência digna de nota. & Baixos volumes pulmonares com cicatriz bilateral irregular versus atelectasia. Caso contrário, nenhuma anormalidade pulmonar aguda ou pulmonar. & O relatório indica que o coração possui contornos mediastinais normais. Observou-se a presença de atelectasia subsegmentar ou cicatriz no meio do pulmão direito e no lobo inferior esquerdo. Não foram encontradas doenças focais no espaço aéreo, nem derrame pleural ou pneumotórax. Os volumes pulmonares estão baixos. Quanto às estruturas ósseas visualizadas, nenhuma apresentou características dignas de nota. & Atelectasia subsegmentar versus cicatriz no pulmão direito e no lobo inferior esquerdo. Sem doença focal do espaço aéreo. Baixos volumes pulmonares. \\ \hline

\end{longtable}
\end{center}

\end{landscape}
\end{@twocolumnfalse}
    \textbf{Acknowledgement}\\
    We acknowledge the Plattform für KI-Translation Essen (KITE) project from the REACT-EU initiative (\url{https://kite.ikim.nrw/}, FRE-0801977), and this work was supported by the FCT within the R\&D Units Project Scope: UIDB/00319/2020.

    \vspace{1em}
    
    \textbf{Conflicts of interest}\\
    All authors have read and approved the manuscript. The authors also declare no conflicts of interest regarding the publication of this paper.

	\bibliographystyle{plainnat}
	\bibliography{References}
\end{document}